\pdfoutput=1

\documentclass[10pt,twocolumn,letterpaper]{article}

\usepackage[pagenumbers]{cvpr}

\usepackage{multirow}
\newcommand{\tablefont}{\rmfamily}
\usepackage{tabularx}
\usepackage{colortbl}
\usepackage{pifont}
\usepackage{bm}
\usepackage{tikz}
\usetikzlibrary{positioning,arrows.meta,fit,backgrounds,calc,shapes.geometric,decorations.pathreplacing}
\definecolor{oursbg}{HTML}{E8F1FB}
\definecolor{oursc}{HTML}{2A78D6}
\definecolor{ctxc}{HTML}{3D3C39}
\definecolor{pcadc}{HTML}{1BAF7A}
\definecolor{gpac}{HTML}{EB6834}
\newcommand{\cmark}{\ding{51}}

\newcommand{\method}{SPS-GS\xspace}
\newcommand{\pcad}{PCAD\xspace}
\newcommand{\gpa}{GPA\xspace}
\DeclareRobustCommand{\appref}{\textcolor{cvprblue}{Appendix}\xspace}

\DeclareMathOperator*{\argmin}{arg\,min}
\newcommand{\R}{\mathbb{R}}
\newcommand{\cF}{\mathcal{F}}
\newcommand{\cG}{\mathcal{G}}
\newcommand{\cR}{\mathcal{R}}
\newcommand{\cD}{\mathcal{D}}
\newcommand{\cS}{\mathcal{S}}
\newcommand{\cU}{\mathcal{U}}
\newcommand{\cB}{\mathcal{B}}
\newcommand{\bd}{\mathbf{d}}
\newcommand{\bM}{\mathbf{M}}
\newcommand{\bu}{\mathbf{u}}
\newcommand{\br}{\mathbf{r}}
\newcommand{\bmu}{\bm{\mu}}
\newcommand{\bSigma}{\bm{\Sigma}}
\newcommand{\bc}{\mathbf{c}}
\newcommand{\bx}{\mathbf{x}}
\newcommand{\by}{\mathbf{y}}

\usepackage{algorithm}
\usepackage{algpseudocode}
\algrenewcommand\algorithmicrequire{\textbf{Input:}}
\algrenewcommand\algorithmicensure{\textbf{Output:}}
\algrenewcommand\algorithmicdo{\textbf{do:}}

\AtEndPreamble{\crefname{algorithm}{Alg.}{Algs.}\Crefname{algorithm}{Algorithm}{Algorithms}}

\definecolor{cvprblue}{rgb}{0.21,0.49,0.74}
\usepackage[pagebackref,breaklinks,colorlinks,allcolors=cvprblue]{hyperref}

\title{You Only Encode Once: Sub-Pixel-Shifted Frames for Arbitrary-Scale\\ Remote Sensing Super-Resolution with 2D Gaussian Splatting}

\author{%
\textbf{Canjie Liu}\textsuperscript{1,*}, 
\textbf{Liuqian Wang}\textsuperscript{1,*,\textdagger}, 
\textbf{Qi Ming}\textsuperscript{2}, 
\textbf{Wei Lu}\textsuperscript{3}\\[3pt]
\textbf{Xiaohan Hao}\textsuperscript{1}, 
\textbf{Yuefei Wu}\textsuperscript{1}, 
\textbf{Wentao Chao}\textsuperscript{1,\textdaggerdbl}, 
\textbf{Lei Shi}\textsuperscript{1}\\[3pt]
{\small\textsuperscript{1}Zhengzhou University\qquad}
{\small\textsuperscript{2}Beijing University of Technology\qquad
\textsuperscript{3}Anhui University}\\
{\footnotesize\textsuperscript{*}Equal contribution\qquad
\textsuperscript{\textdagger}Project Lead\qquad
\textsuperscript{\textdaggerdbl}Corresponding author}
}

\begin{document}
\maketitle

\begin{abstract}
Arbitrary-scale super-resolution (ASSR) of remote sensing images remains challenging due to the limited high-frequency information in a single low-resolution (LR) observation.
While multi-frame SR can exploit complementary sub-pixel information, existing approaches often require real multi-frame training sequences and repeated feature encoding, increasing data acquisition demands and computational costs.
To address these limitations, we propose SPS-GS (\textbf{S}ub-\textbf{P}ixel-\textbf{S}hifted \textbf{G}aussian \textbf{S}platting), an encode-once framework that learns from multi-phase LR observations synthesized from individual high-resolution images, without requiring real multi-frame training sequences. Specifically, Phase-Coherent Aliasing Demodulation (PCAD) exploits shift-dependent aliasing through Fourier-based alignment, phase-aware residual fusion, and least-squares aliasing estimation, enabling multi-frame information to be integrated before a single deep encoding pass. Gaussian Projection Alignment (GPA) then iteratively refines Gaussian centers and colors using multi-observation back-projection to promote observation consistency. The refined Gaussians support continuous rendering at arbitrary scales. Extensive experiments on five remote sensing datasets demonstrate strong reconstruction performance against single-image and multi-frame SR baselines, while maintaining competitive inference efficiency at common scales. The code will be publicly released.

\end{abstract}

\begin{figure}[t]
  \centering
  \includegraphics[width=\linewidth]{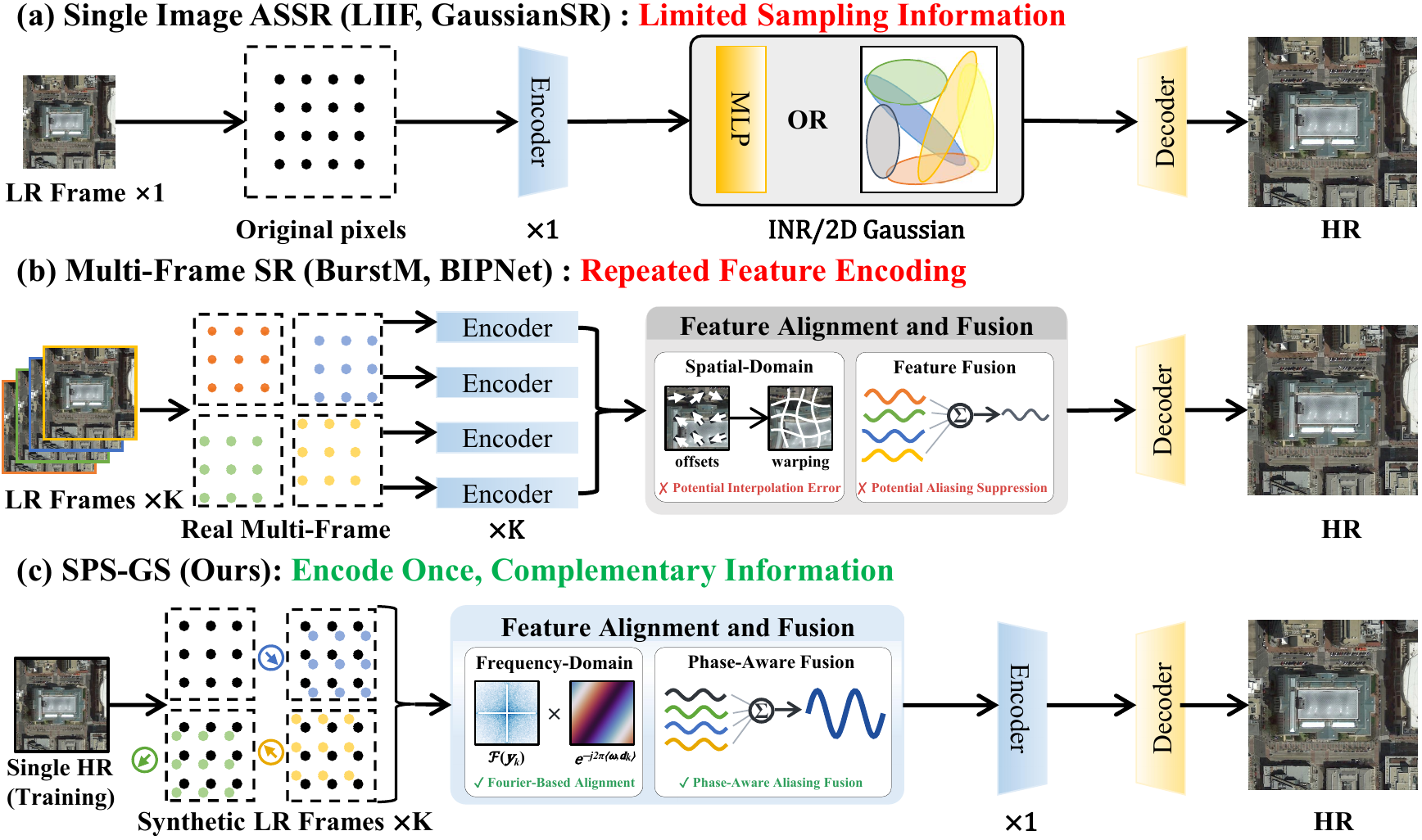}
  \vspace{-1.5em}

    \caption{\textbf{Motivation and comparison of three SR paradigms.}
(a) Single-image ASSR is limited by the sampling information available from one LR observation.
(b) Conventional multi-frame SR exploits complementary observations but often involves repeated feature encoding and spatial-domain alignment, which may suppress aliasing information.
(c) SPS-GS synthesizes complementary LR frames from a single HR image during training and performs phase-aware fusion before a single deep encoding pass at inference.}

  \label{fig:teaser}
  \vspace{-0.6em}
\end{figure}
\section{Introduction}
\label{sec:intro}

Remote sensing images provide essential visual information for Earth observation, environmental monitoring, and geographic analysis~\cite{cheng2017remote,Wang2023FilterGrafting,Wang2025PGDRFNet,Wang2026WeakSupervision,Ming2026HERO,Gan2026GAV}. However, their spatial resolution is constrained by imaging sensors and acquisition conditions, while downstream applications often require imagery at different spatial scales. Arbitrary-scale super-resolution (ASSR) addresses this demand by reconstructing high-resolution (HR) images at flexible magnification factors from low-resolution (LR) observations~\cite{fiete1999image,lei2021hybrid,xiao2024ttst,fmsr2025frequency,Wang2026MOSAIC}. Despite recent advances, recovering fine structures remains challenging because LR sampling discards high-frequency information. With only a single LR observation, reconstruction is inherently underconstrained and relies heavily on learned image priors.

Existing ASSR methods employ implicit neural representations (INRs)~\cite{chen2021learning,lee2022local,cao2023ciaosr,hiif2025hierarchical}
or 2D Gaussian splatting (2DGS)~\cite{hu2025gaussiansr,chen2025generalized,peng2025pixel,Dong2026CPGS}
for continuous reconstruction from a single LR observation (\cref{fig:teaser}(a)), but remain limited by the insufficient high-frequency information in a single LR observation.
Alternatively, multi-frame SR exploits complementary observations of the same scene~\cite{deudon2020highres,bhat2021deep,kang2024burstm},
yet often requires repeated feature encoding followed by alignment and fusion, introducing additional computational overhead (\cref{fig:teaser}(b)).
These limitations motivate an efficient framework that exploits complementary multi-frame information through a single deep encoding pass.

Our key insight is that complementary sub-pixel sampling information can be aggregated before deep feature encoding, allowing multiple LR observations to share a single encoding pass.
This motivates three key designs.
\textbf{(i) Single-source, encode-once reconstruction.}
Multiple complementary LR observations are synthesized from a single HR image through sub-pixel-shifted sampling during training, eliminating the need for real multi-frame training sequences. At inference, multiple LR observations are fused before a single deep encoding pass, as illustrated in \cref{fig:teaser}(c).
\textbf{(ii) Phase-coherent information recovery.}
Sub-pixel shifts introduce distinct phase modulations to aliased frequency components, whose complementary information may be suppressed by phase-agnostic fusion, motivating phase-aware aggregation.
\textbf{(iii) Observation-consistent Gaussian reconstruction.}
Feature fusion alone does not ensure consistency with the observed LR frames. The Gaussian representation is therefore refined using multi-observation reconstruction errors to promote observation consistency while supporting continuous rendering at arbitrary scales.

Motivated by these insights, we propose \textbf{SPS-GS} (\textbf{S}ub-\textbf{P}ixel-\textbf{S}hifted \textbf{G}aussian \textbf{S}platting), an encode-once framework for arbitrary-scale remote sensing image super-resolution based on 2D Gaussian splatting.
Specifically, Phase-Coherent Aliasing Demodulation (\pcad) combines Fourier-based alignment, phase-aware residual fusion, and least-squares aliasing estimation to exploit complementary aliasing information before a single deep encoding pass.
Gaussian Projection Alignment (\gpa) then iteratively refines Gaussian colors and centers using multi-observation back-projection to promote consistency with the observed LR frames.
The resulting Gaussian representation supports efficient rendering at arbitrary scales.
Extensive experiments on five remote sensing datasets demonstrate strong reconstruction performance against single-image and multi-frame SR baselines, while maintaining competitive inference efficiency at common scales.

Our main contributions are as follows:
\begin{itemize}[leftmargin=*,itemsep=1pt,topsep=2pt]
  \item We propose \method, an encode-once framework for ASSR that integrates complementary sub-pixel-shifted LR observations into a unified 2D Gaussian representation, enabling single-source synthetic training and multi-frame reconstruction with a single deep encoding pass.
  \item We develop Phase-Coherent Aliasing Demodulation (\pcad) to exploit shift-dependent aliasing information through Fourier alignment, phase-aware fusion, and least-squares estimation before deep encoding.
  \item We introduce Gaussian Projection Alignment (\gpa), which iteratively refines Gaussian colors and centers through multi-observation back-projection to promote consistency with the observed LR frames while supporting arbitrary-scale rendering.
  \item Extensive experiments on five remote sensing datasets validate the effectiveness and efficiency of SPS-GS across both seen and unseen super-resolution scales.
\end{itemize}

\begin{figure*}[!t]
  \centering
  \includegraphics[width=\textwidth]{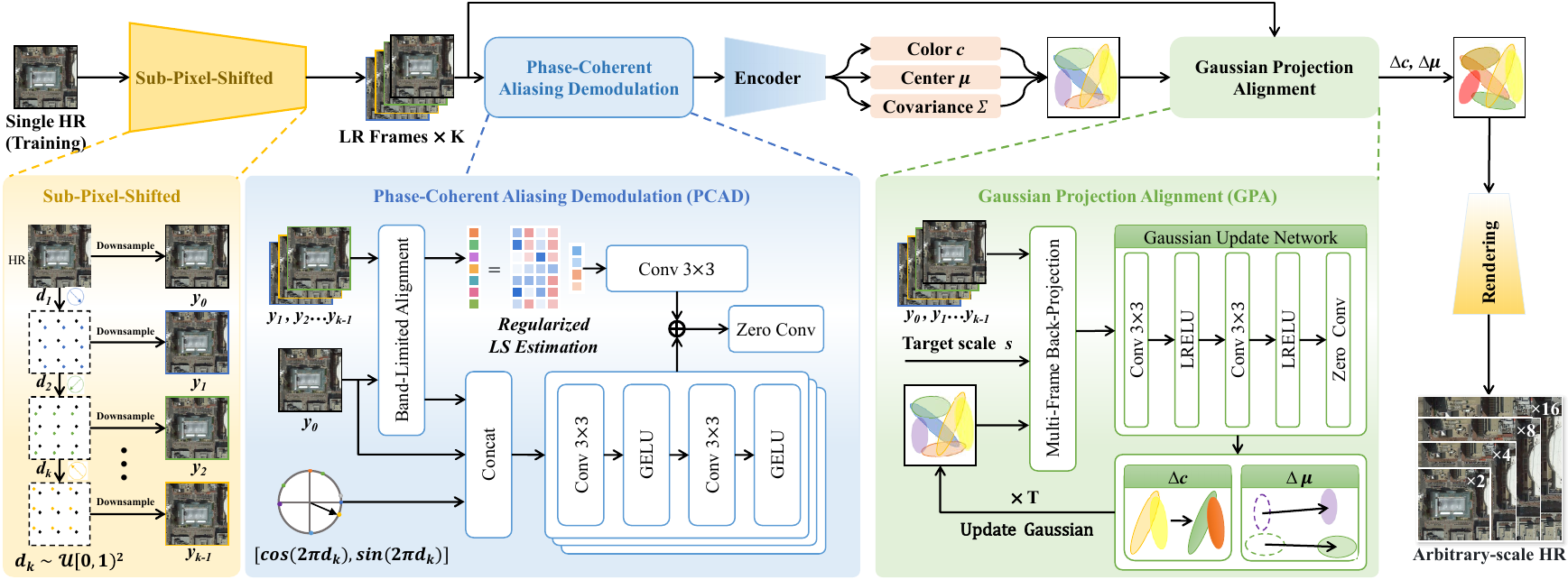}
  \vspace{-1.6em}
  \caption{\textbf{Overview of SPS-GS.}
During training, multiple sub-pixel-shifted LR frames are synthesized from a single HR image (left).
PCAD (middle) integrates complementary aliasing information through Fourier alignment, phase-aware fusion, and regularized least-squares estimation before a single deep encoding pass.
GPA (right) iteratively refines Gaussian colors and centers through multi-frame back-projection, enabling arbitrary-scale rendering.
At inference, the framework directly takes multiple LR observations as input.}
  \label{fig:framework}
  \vspace{-0.6em}
\end{figure*}

\section{Related Work}
\label{sec:related}

\subsection{Remote Sensing Super-Resolution}

Remote sensing SR has evolved from CNNs~\cite{lei2021hybrid} to transformers~\cite{xiao2024ttst}, diffusion models~\cite{xiao2024ediffsr}, and state space models~\cite{fmsr2025frequency}, mainly targeting fixed scales.
ASSR enables flexible reconstruction through scale-conditioned filters~\cite{hu2019meta} and implicit functions~\cite{chen2021learning}, further advanced by texture modeling~\cite{lee2022local}, attention~\cite{cao2023ciaosr}, neural operators~\cite{wei2023super}, latent modulation~\cite{he2024latent}, neural heat fields~\cite{becker2025thera}, and hierarchical encoding~\cite{hiif2025hierarchical}.
Remote sensing ASSR explores implicit-space interaction~\cite{chen2023continuous}, scale awareness~\cite{wu2023learning}, neural-operator diffusion~\cite{neuropdiff2025}, and Gaussian reconstruction~\cite{zhou2026tadsr,mo2026flowgs}.
Unlike these predominantly single-image methods, \method exploits complementary sub-pixel-shifted observations through pre-encoding fusion for arbitrary-scale reconstruction.

\subsection{2D Gaussian Splatting for Super-Resolution}
Gaussian splatting represents visual content using anisotropic Gaussian primitives and efficient rasterization~\cite{kerbl20233d}.
GaussianImage~\cite{zhang2024gaussianimage}, Image-GS~\cite{zhang2025image}, and LIG~\cite{zhu2025large} optimize Gaussian primitives for image representation and compression.
For ASSR, feed-forward networks instead predict Gaussian representations.
GaussianSR~\cite{hu2025gaussiansr} decodes rendered Gaussian features into colors, while GSASR~\cite{chen2025generalized} predicts image-adaptive Gaussians for direct arbitrary-scale rendering.
ContinuousSR~\cite{peng2025pixel} employs pixel-wise Gaussians with a learned covariance dictionary for efficient reconstruction.
TADSR~\cite{zhou2026tadsr} and FlowGS~\cite{mo2026flowgs} extend Gaussian-based SR to remote sensing imagery.
Unlike these predominantly single-image methods, \method combines feed-forward Gaussian prediction with multi-observation refinement, where \gpa iteratively updates Gaussian centers and colors using LR reconstruction errors to promote observation consistency.

\subsection{Multi-Frame Super-Resolution}
Classical multi-frame SR exploits complementary sub-pixel samples through generalized sampling theory~\cite{papoulis1977generalized} and frequency-domain reconstruction~\cite{tsai1984multiframe}, followed by iterative back-projection~\cite{irani1991improving} and joint registration and reconstruction~\cite{hardie1997joint}.
These principles have been applied to satellite imaging systems, including SPOT-5~\cite{latry2003super} and SkySat~\cite{lafenetre2023handheld}.
Deep multi-frame SR has advanced through PROBA-V reconstruction~\cite{deudon2020highres,an2022tr,valsesia2022permutation}, multi-temporal and cross-sensor fusion~\cite{cornebise2022open,tarasiewicz2025graph,retnanto2026beyond}, and satellite video SR~\cite{xiao2024local,ni2024deformable,ni2026satellite}.
Related burst SR methods exploit frame alignment and feature fusion\cite{bhat2021deep,wei2023towards,dudhane2025burst,di2025qmambabsr,kang2024burstm}.
Meanwhile, deep back-projection networks~\cite{haris2018deep,zhang2020deep} and diffusion-based restoration~\cite{kawar2022denoising} incorporate observation-driven reconstruction.
Building on these principles, \method integrates shift-dependent aliasing information before a single deep encoding pass and refines Gaussian parameters through multi-observation back-projection.

\section{Preliminaries}
\label{sec:prelim}

\paragraph{2D Gaussian representation.}
Following ContinuousSR~\cite{peng2025pixel}, an HR image is represented by a set $\cG=\{(\bmu_i,\bSigma_i,\bc_i)\}_{i=1}^{N}$ of $N$ 2D Gaussians, where $\bmu_i\in\R^2$, $\bSigma_i\in\R^{2\times2}$, and $\bc_i\in\R^3$ denote the center, covariance matrix, and RGB color of the $i$-th Gaussian, respectively.
Given a target scale factor $s$, the image is reconstructed through Gaussian rasterization:
\begin{equation}
  \cR_s(\cG)(\mathbf{p})=\sum_{i=1}^{N} \bc_i\exp\!\Big(-\tfrac12(\mathbf{p}-\bmu_i)^{\!\top}\bSigma_i^{-1}(\mathbf{p}-\bmu_i)\Big),
  \label{eq:render}
\end{equation}
where $\mathbf{p}\in\R^2$ denotes an output pixel center on the scale-dependent sampling grid, and rasterization is implemented using tile-based splatting~\cite{kerbl20233d,zhang2024gaussianimage}.
The target scale $s$ determines the sampling grid, allowing the same Gaussian primitives to be rendered at different resolutions without scale-specific parameterization.

\section{Method}
\label{sec:method}

\subsection{Overview}
\label{sec:overview}

\Cref{fig:framework} summarizes \method (pseudocode in the \appref).
During training, we synthesize $K$ sub-pixel-shifted LR observations from a single HR image (\cref{sec:frames}), while inference takes $K$ LR observations as input.
\pcad (\cref{sec:pcad}) aggregates complementary shift-dependent aliasing information relative to a reference frame before a single deep encoding pass, producing features $\mathbf{f}$.
Lightweight prediction heads then initialize $\cG^{(0)}$ with one Gaussian per feature location, following ContinuousSR~\cite{peng2025pixel}.
Finally, \gpa (\cref{sec:gpa}) refines Gaussian colors and centers over $T$ learned iterations guided by multi-frame back-projection errors, and the resulting representation is rasterized at arbitrary scales using \cref{eq:render}.

\subsection{Sub-Pixel-Shifted Frames}
\label{sec:frames}
Let $\bx\in\R^{3\times sH\times sW}$ denote an HR image. During training, we synthesize $K$ LR observations $\by_k\in\R^{3\times H\times W}$, $k=0,\ldots,K-1$, by sampling the same image at different sub-pixel positions:
\begin{equation}
  \by_k=\cD_s\big(\cS_{\bd_k}\bx\big)=A_k\bx,
  \label{eq:obs}
\end{equation}
where $\bd_k\in[0,1)^2$ denotes the shift in LR pixel units, $\cS_{\bd_k}$ translates the HR image by $s\bd_k$ pixels using bilinear interpolation, and $\cD_s$ denotes anti-aliased bicubic downsampling by a factor of $s$. The resulting observation operator is $A_k=\cD_s\cS_{\bd_k}$. We set $\bd_0=\mathbf{0}$ as the reference and sample the remaining shifts uniformly from $[0,1)^2$. This single-source synthesis eliminates the need for real multi-frame training sequences while approximating the sub-pixel sampling diversity encountered in satellite imaging~\cite{latry2003super,lafenetre2023handheld}.

While all observations originate from the same HR image, their shifted sampling grids provide complementary measurements. Under the linear observation model, the ambiguity of a single observation is characterized by $\mathcal{N}(A_0)$, whereas the joint ambiguity is characterized by $\bigcap_{k=0}^{K-1}\mathcal{N}(A_k)\subseteq\mathcal{N}(A_0)$. This motivates phase-aware fusion in \pcad (\cref{sec:pcad}) and multi-frame consistency refinement in \gpa (\cref{sec:gpa}).

\subsection{Phase-Coherent Aliasing Demodulation}
\label{sec:pcad}
\pcad exploits complementary aliasing information across the $K$ LR observations and integrates it with the reference frame before a single deep encoding pass.

\paragraph{Band-Limited Sub-Pixel Alignment.}
\label{sec:align}
Each non-reference frame $\by_k$, $k=1,\dots,K-1$, is approximately aligned to the reference grid using Fourier-domain phase modulation. The residual relative to the reference frame is computed as
\begin{equation}
  \tilde\by_k=\cF^{-1}\big[\cF[\by_k]\,e^{-\jmath2\pi\langle\bm{\omega},\bd_k\rangle}\big],\qquad
  \br_k=\tilde\by_k-\by_0,
  \label{eq:align}
\end{equation}
where $\tilde\by_k$ denotes the aligned frame, $\cF$ and $\cF^{-1}$ are the 2D discrete Fourier transform and its inverse, $\bm{\omega}\in[-\frac12,\frac12)^2$ denotes spatial frequency in cycles per LR pixel, and $\jmath$ is the imaginary unit. Under an idealized sampling model, Fourier alignment compensates for the baseband shift, while aliased components retain shift-dependent relative phases. In practice, interpolation and boundary effects may introduce discrepancies, so the residual $\br_k$ approximately captures complementary aliasing information from frame $k$. Using a truncated model of low-order aliasing components, the residual can be approximated as
\begin{equation}
  \br_k\approx\sum_{\mathbf{m}\in\cB}\big[(\cos\theta_{k,\mathbf{m}}-1)\,\mathbf{a}_{\mathbf{m}}+\sin\theta_{k,\mathbf{m}}\,\mathbf{b}_{\mathbf{m}}\big],
  \label{eq:alias}
\end{equation}
where $\theta_{k,\mathbf{m}}=2\pi\langle\mathbf{m},\bd_k\rangle$ is the shift-dependent phase (\cref{sec:prelim}), $\mathbf{a}_{\mathbf{m}}$ and $\mathbf{b}_{\mathbf{m}}$ denote the in-phase and quadrature components, and $\cB=\{(1,0),(0,1),(1,1),(1,-1)\}$ specifies the selected low-order aliasing orders (derivation in the \appref). Their shift-dependent coefficients may change sign across frames, motivating phase-aware aggregation rather than naive residual summation.

\paragraph{Phase-Aware Branch.}
\label{sec:branch}
A lightweight convolutional branch $\psi$, shared across frames, processes each residual together with its sub-pixel shift:
\begin{equation}
  \mathbf{z}=\sum_{k=1}^{K-1}\psi\big([\by_0,\,\br_k,\,\phi(\bd_k)]\big),
  \label{eq:branch}
\end{equation}
where $[\cdot]$ denotes channel-wise concatenation, $\phi(\bd)=[\cos(2\pi\bd),\sin(2\pi\bd)]$ is a four-dimensional shift encoding broadcast across spatial locations, and $\mathbf{z}$ denotes the aggregated feature map. This encoding provides shift-dependent phase cues, enabling $\psi$ to adaptively combine residuals with different signs and phases. The shared branch and summation make $\mathbf{z}$ invariant to permutations of the non-reference frames.

\paragraph{Regularized Least-Squares Aliasing Estimation.}
\label{sec:demod}
In parallel, \pcad estimates selected low-order aliasing components independently at each pixel and color channel through regularized least squares. We stack the $K-1$ residual values into $\br\in\R^{K-1}$ and construct $\bM\in\R^{(K-1)\times2|\cB_{\mathrm{LS}}|}$, whose rows contain $(\cos\theta_{k,\mathbf{m}}-1,\,\sin\theta_{k,\mathbf{m}})_{\mathbf{m}\in\cB_{\mathrm{LS}}}$. The estimated components are given by
\begin{equation}
  \bu=\big(\bM^{\!\top}\bM+\lambda\mathbf{I}\big)^{-1}\bM^{\!\top}\br,
  \label{eq:demod}
\end{equation}
where $\lambda>0$ is the Tikhonov regularization parameter and $\bu\in\R^{2|\cB_{\mathrm{LS}}|}$ contains the estimated in-phase and quadrature components. We select $\cB_{\mathrm{LS}}=\{(1,0),(0,1)\}$, yielding four unknowns from $K-1=7$ residual observations. Under the truncated aliasing model, the unregularized solution is unique when $\bM$ has full column rank, while regularization improves numerical stability for poorly conditioned shifts. Including all four aliasing orders in \cref{eq:alias} would introduce eight unknowns, making the unregularized system underdetermined for $K=8$. Thus, this parameter-free estimation captures selected low-order components rather than the complete aliasing spectrum. Finally, the reference frame, aligned residuals, $\mathbf{z}$, and $\bu$ are fused through zero-initialized convolutions before a single deep encoding pass.

\setcounter{dbltopnumber}{2}
\begin{table*}[t]
\centering
\caption{PSNR (dB); every dataset has its own training and test sets. $\times2$ and $\times4$ are seen scales, $\times8$ and $\times16$ unseen. \textbf{Bold}: best; \underline{underline}: second best. All scales in the \appref.}
\label{tab:main}
\tablefont
\setlength{\tabcolsep}{2.4pt}
\resizebox{\textwidth}{!}{%
\begin{tabular}{l|cccc|cccc|cccc|cccc|cccc}
\toprule
\multirow{2}{*}{Methods} & \multicolumn{4}{c|}{AID~\cite{xia2017aid}} & \multicolumn{4}{c|}{UCMerced~\cite{yang2010bag}} & \multicolumn{4}{c|}{WHU-RS19~\cite{dai2011satellite}} & \multicolumn{4}{c|}{RSSCN7~\cite{zou2015deep}} & \multicolumn{4}{c}{NWPU-RESISC45~\cite{cheng2017remote}} \\
 & $\times2$ & $\times4$ & $\times8$ & $\times16$ & $\times2$ & $\times4$ & $\times8$ & $\times16$ & $\times2$ & $\times4$ & $\times8$ & $\times16$ & $\times2$ & $\times4$ & $\times8$ & $\times16$ & $\times2$ & $\times4$ & $\times8$ & $\times16$ \\
\midrule
Bicubic & 32.90 & 27.58 & 24.56 & 22.60 & 31.09 & 25.70 & 22.48 & 20.42 & 33.67 & 27.73 & 24.45 & 22.42 & 30.24 & 26.36 & 24.10 & 22.55 & 30.98 & 26.36 & 23.66 & 21.77 \\
\midrule
\multicolumn{21}{c}{\emph{Single-image methods}} \\
Meta-SR~\cite{hu2019meta} & 35.75 & 29.60 & 25.84 & 23.44 & 35.10 & 28.60 & 24.17 & 21.31 & 36.65 & 29.83 & 25.78 & 23.23 & 32.53 & 27.80 & 25.03 & 23.15 & 33.51 & 28.15 & 24.76 & 22.43 \\
LIIF~\cite{chen2021learning} & 35.75 & 29.66 & 25.91 & 23.51 & 35.16 & 28.74 & 24.33 & 21.42 & 36.72 & 29.92 & 25.88 & 23.35 & 32.53 & 27.85 & 25.10 & 23.25 & 33.55 & 28.19 & 24.84 & 22.54 \\
LTE~\cite{lee2022local} & 35.77 & 29.68 & 25.05 & 21.54 & 35.17 & 28.74 & 23.80 & 19.98 & 36.74 & 29.90 & 24.96 & 21.54 & 32.54 & 27.86 & 24.11 & 20.49 & 33.58 & 28.22 & 24.30 & 21.06 \\
LMF~\cite{he2024latent} & 35.76 & 29.67 & 25.86 & 23.41 & 35.15 & 28.75 & 24.28 & 21.29 & 36.68 & 29.90 & 25.15 & 22.16 & 32.53 & 27.86 & 24.96 & 23.00 & 33.56 & 28.20 & 24.64 & 22.21 \\
HIIF~\cite{hiif2025hierarchical} & 35.77 & 29.68 & 25.93 & 23.49 & 35.16 & 28.73 & 24.33 & 21.42 & 36.71 & 29.92 & 25.86 & 23.32 & 32.55 & 27.87 & 25.10 & 23.21 & 33.57 & 28.21 & 24.84 & 22.51 \\
NeurOp-Diff~\cite{neuropdiff2025} & 30.43 & 28.13 & 25.28 & 23.11 & 29.88 & 26.50 & 23.18 & 20.78 & 30.08 & 27.34 & 24.50 & 22.39 & 27.79 & 26.22 & 24.35 & 22.73 & 29.05 & 26.76 & 24.25 & 22.16 \\
GaussianSR~\cite{hu2025gaussiansr} & 35.76 & 29.66 & 25.77 & 23.22 & 35.08 & 28.64 & 24.12 & 21.09 & 36.68 & 29.91 & 25.62 & 22.90 & 32.52 & 27.83 & 24.96 & 22.97 & 33.55 & 28.20 & 24.70 & 22.27 \\
GSASR~\cite{chen2025generalized} & 35.72 & 29.63 & 25.91 & 23.50 & 34.99 & 28.52 & 24.21 & 21.33 & 36.54 & 29.77 & 25.80 & 23.30 & 32.47 & 27.81 & 25.08 & 23.23 & 33.52 & 28.16 & 24.82 & 22.50 \\
ContinuousSR~\cite{peng2025pixel} & 35.69 & 29.67 & 25.92 & 23.51 & 35.10 & 28.70 & 24.33 & 21.43 & 36.56 & 29.85 & 25.83 & 23.33 & 32.52 & 27.84 & 25.10 & 23.25 & 33.23 & 28.17 & 24.82 & 22.52 \\
\midrule
\multicolumn{21}{c}{\emph{Multi-frame methods}} \\
BurstM~\cite{kang2024burstm} & 37.52 & 30.78 & 24.27 & 21.88 & 36.71 & 30.60 & 22.39 & 19.94 & 39.07 & 31.56 & 24.00 & 21.58 & 34.58 & 29.10 & 23.91 & 21.93 & 36.17 & 30.08 & 23.39 & 21.20 \\
BIPNet~\cite{dudhane2025burst} & 38.07 & 31.09 & 26.59 & 23.86 & 36.88 & 29.82 & 24.89 & 21.05 & 38.67 & 30.89 & 26.28 & 23.42 & 35.58 & 28.73 & 25.43 & 23.00 & 35.31 & 29.47 & 25.40 & 22.32 \\
HighRes-net~\cite{deudon2020highres} & 38.10 & 30.77 & 26.44 & \underline{23.86} & 36.53 & 29.16 & 24.64 & \underline{21.75} & 39.53 & 31.04 & 26.31 & \underline{23.57} & 35.25 & 28.74 & 25.59 & \underline{23.60} & 36.24 & 29.28 & 25.39 & \underline{22.97} \\
TR-MISR~\cite{an2022tr} & 37.62 & 29.93 & 26.26 & 23.71 & 35.88 & 28.35 & 24.41 & 21.58 & 38.67 & 30.09 & 26.06 & 23.39 & 34.75 & 28.08 & 25.44 & 23.46 & 35.63 & 28.47 & 25.19 & 22.79 \\
PIUnet~\cite{valsesia2022permutation} & 29.44 & 28.05 & 26.12 & 23.80 & 32.42 & 29.16 & 24.62 & 21.39 & 30.68 & 28.71 & 26.03 & 23.50 & 30.00 & 27.88 & 25.48 & 23.34 & 31.92 & 28.88 & 25.25 & 22.51 \\
ST-AVSR~\cite{shang2024arbitrary} & \underline{39.87} & \underline{32.27} & \underline{26.76} & 23.61 & \underline{38.58} & \underline{30.86} & \underline{24.94} & 21.49 & \underline{41.77} & \underline{32.86} & \underline{26.72} & 23.44 & \underline{37.25} & \underline{29.79} & \underline{25.75} & 23.33 & \underline{38.37} & \underline{30.46} & \underline{25.60} & 22.67 \\
\midrule
SPS-GS (Ours) & \textbf{40.60} & \textbf{32.72} & \textbf{27.52} & \textbf{24.40} & \textbf{40.02} & \textbf{31.78} & \textbf{26.15} & \textbf{22.60} & \textbf{42.30} & \textbf{33.22} & \textbf{27.43} & \textbf{24.01} & \textbf{37.94} & \textbf{30.13} & \textbf{26.32} & \textbf{23.90} & \textbf{38.77} & \textbf{31.03} & \textbf{26.45} & \textbf{23.59} \\
\bottomrule
\end{tabular}}
\end{table*}

\begin{table*}[t]
\centering
\caption{SSIM, LPIPS, DISTS and FID on AID (all scales in the \appref). \textbf{Bold}: best; \underline{underline}: second best.}
\label{tab:metrics}
\tablefont
\setlength{\tabcolsep}{2.6pt}
\resizebox{\textwidth}{!}{%
\begin{tabular}{l|cccc|cccc|cccc|cccc}
\toprule
\multirow{2}{*}{Methods} & \multicolumn{4}{c|}{SSIM$\uparrow$} & \multicolumn{4}{c|}{LPIPS$\downarrow$} & \multicolumn{4}{c|}{DISTS$\downarrow$} & \multicolumn{4}{c}{FID$\downarrow$} \\
 & $\times2$ & $\times4$ & $\times8$ & $\times16$ & $\times2$ & $\times4$ & $\times8$ & $\times16$ & $\times2$ & $\times4$ & $\times8$ & $\times16$ & $\times2$ & $\times4$ & $\times8$ & $\times16$ \\
\midrule
Bicubic & 0.9083 & 0.7405 & 0.6075 & 0.5538 & 0.2133 & 0.4912 & 0.7125 & 0.8514 & 0.1081 & 0.2185 & 0.3400 & 0.4839 & 14.92 & 39.94 & 98.80 & 208.60 \\
\midrule
\multicolumn{17}{c}{\emph{Single-image methods}} \\
Meta-SR~\cite{hu2019meta} & 0.9466 & 0.8169 & 0.6670 & 0.5752 & 0.1059 & 0.3046 & 0.5101 & 0.7064 & 0.0757 & 0.1680 & 0.2656 & 0.3725 & 9.72 & 43.04 & 94.87 & 168.22 \\
LIIF~\cite{chen2021learning} & 0.9467 & 0.8171 & 0.6702 & 0.5799 & 0.1043 & 0.3186 & 0.5360 & 0.7277 & 0.0752 & 0.1736 & 0.2745 & 0.3850 & 9.73 & 42.27 & 97.22 & 166.94 \\
LTE~\cite{lee2022local} & 0.9469 & 0.8177 & 0.6668 & 0.5741 & 0.1049 & 0.3149 & 0.5324 & 0.7275 & 0.0759 & 0.1739 & 0.2722 & 0.3855 & 9.65 & 43.30 & 86.86 & 151.63 \\
LMF~\cite{he2024latent} & 0.9470 & 0.8173 & 0.6689 & 0.5753 & 0.1045 & 0.3166 & 0.5334 & 0.7233 & 0.0759 & 0.1738 & 0.2704 & 0.3725 & 9.53 & 43.03 & 86.63 & 149.63 \\
HIIF~\cite{hiif2025hierarchical} & 0.9468 & 0.8178 & 0.6729 & 0.5816 & 0.1059 & 0.3150 & 0.5258 & 0.7143 & 0.0761 & 0.1740 & 0.2719 & 0.3737 & 9.49 & 43.41 & 98.67 & 156.31 \\
NeurOp-Diff~\cite{neuropdiff2025} & 0.8763 & 0.7733 & 0.6389 & 0.5581 & 0.1298 & 0.2628 & 0.4888 & 0.7150 & 0.0957 & 0.1501 & 0.2725 & 0.3944 & 11.87 & 24.53 & 83.96 & 148.12 \\
GaussianSR~\cite{hu2025gaussiansr} & 0.9465 & 0.8170 & 0.6622 & 0.5710 & 0.1095 & 0.3174 & 0.5497 & 0.7733 & 0.0764 & 0.1719 & 0.2684 & 0.3925 & 9.50 & 41.37 & 78.28 & 146.77 \\
GSASR~\cite{chen2025generalized} & 0.9462 & 0.8161 & 0.6702 & 0.5819 & 0.1102 & 0.3195 & 0.5224 & \underline{0.7037} & 0.0767 & 0.1751 & 0.2750 & 0.3861 & 9.63 & 43.05 & 104.15 & 173.40 \\
ContinuousSR~\cite{peng2025pixel} & 0.9467 & 0.8174 & 0.6700 & 0.5798 & 0.1075 & 0.3181 & 0.5407 & 0.7316 & 0.0755 & 0.1723 & 0.2714 & 0.3825 & 9.52 & 41.92 & 91.09 & 154.93 \\
\midrule
\multicolumn{17}{c}{\emph{Multi-frame methods}} \\
BurstM~\cite{kang2024burstm} & 0.9685 & 0.8627 & 0.5894 & 0.5135 & 0.0235 & 0.2424 & 0.6075 & 0.7785 & 0.0358 & 0.1298 & 0.3817 & 0.5512 & 1.24 & 18.37 & 100.64 & 228.85 \\
BIPNet~\cite{dudhane2025burst} & 0.9752 & 0.8694 & 0.6975 & \underline{0.5862} & 0.0271 & 0.2801 & 0.5346 & 0.7564 & 0.0330 & 0.1275 & 0.2497 & \underline{0.3635} & 1.27 & \textbf{15.64} & 66.70 & \underline{136.41} \\
HighRes-net~\cite{deudon2020highres} & 0.9707 & 0.8541 & 0.6858 & 0.5844 & 0.0380 & 0.3000 & 0.5516 & 0.7703 & 0.0429 & 0.1429 & 0.2573 & 0.3782 & 2.79 & 20.68 & 72.08 & 142.28 \\
TR-MISR~\cite{an2022tr} & 0.9671 & 0.8323 & 0.6785 & 0.5804 & 0.0570 & 0.3137 & 0.5668 & 0.7777 & 0.0520 & 0.1531 & 0.2670 & 0.3946 & 4.32 & 24.88 & 78.53 & 156.00 \\
PIUnet~\cite{valsesia2022permutation} & 0.8952 & 0.8302 & 0.6891 & 0.5844 & 0.1603 & 0.3329 & 0.5445 & 0.7689 & 0.1040 & 0.1600 & 0.2560 & 0.3801 & 13.65 & 28.13 & 81.87 & 140.05 \\
ST-AVSR~\cite{shang2024arbitrary} & \underline{0.9804} & \underline{0.8906} & \underline{0.7009} & 0.5776 & \underline{0.0165} & \underline{0.2340} & \textbf{0.4424} & 0.7175 & \underline{0.0317} & \underline{0.1240} & \underline{0.2260} & 0.3761 & \underline{1.16} & 18.16 & \textbf{53.82} & 136.68 \\
\midrule
SPS-GS (Ours) & \textbf{0.9829} & \textbf{0.8995} & \textbf{0.7372} & \textbf{0.6078} & \textbf{0.0101} & \textbf{0.2127} & \underline{0.4561} & \textbf{0.6996} & \textbf{0.0248} & \textbf{0.1118} & \textbf{0.2211} & \textbf{0.3310} & \textbf{0.82} & \underline{16.86} & \underline{64.05} & \textbf{117.60} \\
\bottomrule
\end{tabular}}
\end{table*}

\subsection{Gaussian Projection Alignment}
\label{sec:gpa}
Given $K$ LR observations of the same scene, \gpa refines a shared Gaussian representation to promote consistency with the observations. The ideal reconstruction objective is formulated as

\begin{equation}
  \hat\bx=\cR_s(\cG^\star),\qquad
  \cG^\star=\argmin_{\cG}\sum_{k=0}^{K-1}\big\|\by_k-A_k\,\cR_s(\cG)\big\|_2^2,
  \label{eq:goal}
\end{equation}
where $\hat\bx$ denotes the reconstructed HR image and $\cG^\star$ the refined Gaussian representation. Rather than solving this objective exactly, \gpa performs a fixed number of learned refinement steps guided by multi-frame reconstruction errors, following the principle of iterative back-projection~\cite{irani1991improving}.

\paragraph{Learned Back-Projection on Gaussian Parameters.}
\label{sec:unfold}
At iteration $t=0,\dots,T-1$, the current Gaussians $\cG^{(t)}$ are rendered and degraded to each LR observation grid. The resulting reconstruction errors are approximately back-projected and aggregated as
\begin{equation}
  \mathbf{e}^{(t)}=\frac1K\sum_{k=0}^{K-1}\cS_{-\bd_k}\,\cU_s\big(\by_k-A_k\,\cR_s(\cG^{(t)})\big),
  \label{eq:bp}
\end{equation}
where $\cU_s$ denotes bicubic upsampling by a factor of $s$, and $\cS_{-\bd_k}$ maps the upsampled errors to the reference HR grid. This approximation aggregates multi-frame reconstruction errors without explicitly computing the adjoint operators $A_k^\top$.

A lightweight convolutional network $\Phi_t$ predicts updates to Gaussian colors and centers:
\begin{equation}
  [\Delta\bc,\Delta\bmu]=\Phi_t\big(\mathbf{f},\,\mathbf{e}^{(t)},\,\mathcal{H}(\mathbf{e}^{(t)}),\,\bc^{(t)},\bmu^{(t)}\big),
  \label{eq:update}
\end{equation}
where $\bc^{(t)}$ and $\bmu^{(t)}$ denote the color and center parameter maps, respectively, and $\mathcal{H}$ extracts the directional detail sub-bands (LH, HL, and HH) using a single-level Haar wavelet transform. These sub-bands provide high-frequency error cues for structural refinement. The Gaussian parameters are updated as $\bc^{(t+1)}=\bc^{(t)}+\Delta\bc$ and $\bmu^{(t+1)}=\bmu^{(t)}+\Delta\bmu$, while the covariances $\bSigma_i$ remain fixed. Center updates allow the representation to adjust the spatial locations of reconstructed structures. After $T$ iterations, the refined Gaussians are rasterized at arbitrary output scales without scale-specific prediction heads.

\begin{figure*}[!t]
  \centering
  \includegraphics[width=\textwidth]{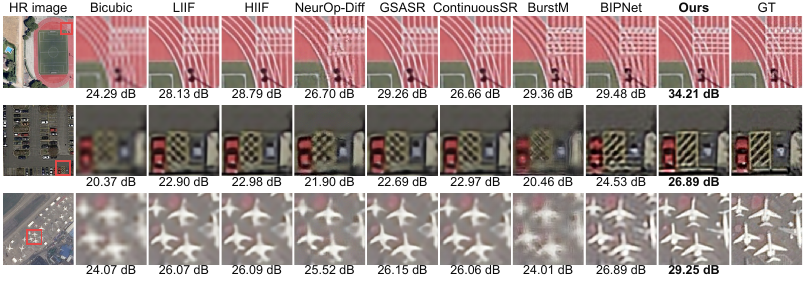}
  \vspace{-1.6em}
  \caption{\textbf{Qualitative comparison on AID at $\times8$ (unseen scale).} Red boxes show zoomed regions; PSNR is measured on full images. Selected high-gain examples are shown here, with random examples in the \appref.}
  \label{fig:qual}
  \vspace{-0.6em}
\end{figure*}

\section{Experiments}
\label{sec:experiments}

\subsection{Experimental Setting}
\label{sec:setup}
\paragraph{Datasets and metrics.}
We evaluate on AID~\cite{xia2017aid}, UCMerced~\cite{yang2010bag}, WHU-RS19~\cite{dai2011satellite}, RSSCN7~\cite{zou2015deep}, and NWPU-RESISC45~\cite{cheng2017remote}, using fixed train/test splits with 2{,}000, 420, 102, 280, and 3{,}150 test images, respectively. For each test image, we synthesize $K{=}8$ LR observations using \cref{eq:obs} with a fixed random seed. Single-image baselines use only the reference frame $y_0$, whereas all multi-frame methods receive the same $K$ LR observations. Additional evaluations on three real-world multi-frame datasets are presented in \cref{sec:realmf}. We report PSNR, SSIM~\cite{wang2004image}, LPIPS~\cite{zhang2018unreasonable}, DISTS~\cite{ding2022image}, and FID~\cite{heusel2017gans}.

\paragraph{Implementation details.}
\method employs a SwinIR encoder~\cite{liang2021swinir} with $K{=}8$ LR observations and $T{=}7$ \gpa iterations. We train using Adam with a learning rate of $10^{-4}$ and a batch size of 32 on a single RTX 4090 GPU. The training loss is $\|\hat\bx-\bx\|_1+0.3\|\cF(\hat\bx)-\cF(\bx)\|_1$, where $\cF$ denotes the 2D FFT. Training scales are uniformly sampled from $s\in[1,4]$, making $\times8$ and $\times16$ unseen during training. On AID, \method is trained for 70k iterations (60k with a constant learning rate followed by 10k with cosine decay). All baselines follow a consistent training and evaluation protocol. Fixed-scale multi-frame SR methods are adapted for arbitrary-scale reconstruction using additional scale-conditioned decoding heads, with implementation details provided in the \appref.

\subsection{Comparison with State-of-the-Art Methods}
\label{sec:comparison}

\paragraph{Quantitative results.}
\Cref{tab:main} shows that \method achieves the highest PSNR across all five datasets and evaluated scales, consistently outperforming both single-image and multi-frame baselines under the same evaluation protocol. Notably, \method maintains its advantage at unseen scales ($\times8$ and $\times16$), demonstrating strong cross-scale generalization.
On AID, it also achieves the best SSIM and DISTS, with competitive LPIPS and FID results (\cref{tab:metrics}; additional results in the \appref).

\begin{table}[t]
\centering

\caption{\textbf{Efficiency comparison on an RTX 4090.} Inference time (ms) at fixed HR resolution ($600\times600$); GPU memory (GiB) at $\times2$, excluding model weights. \textbf{Bold}: best; \underline{underline}: second best.}

\label{tab:efficiency}
\tablefont
\setlength{\tabcolsep}{5.5pt}
\footnotesize
\begin{tabular}{lrrrrr}
\toprule
\multirow{2}{*}{Methods} & \#Params & \multicolumn{3}{c}{Inference time (ms)} & Memory \\
\cmidrule(lr){3-5}
 & (M) & $\times4$ & $\times8$ & $\times16$ & (GiB) \\
\midrule
Meta-SR & 13.68 & 142.5 & 96.9 & 82.5 & 3.81 \\
LIIF & 15.92 & 442.9 & 400.7 & 381.6 & 1.43 \\
LTE & 13.34 & 131.3 & 84.0 & 71.2 & 0.89 \\
LMF & 12.61 & 85.3 & 38.2 & \underline{27.4} & 1.30 \\
HIIF & 14.08 & 244.6 & 200.1 & 182.7 & 6.56 \\
NeurOp-Diff & 49.49 & 3169.4 & 3167.8 & 3163.7 & 3.93 \\
GaussianSR & 12.58 & 109.7 & 66.5 & 56.9 & 4.73 \\
GSASR & 31.19 & 640.7 & 243.4 & 71.5 & 9.73 \\
ContinuousSR & 15.92 & 80.4 & \textbf{21.6} & \textbf{8.0} & 3.44 \\
BurstM & 13.16 & \textbf{71.4} & 57.5 & 55.6 & 1.18 \\
BIPNet & 7.02 & 223.3 & 111.2 & 80.5 & 4.82 \\
HighRes-net & 0.90 & 86.5 & 75.8 & 72.2 & 3.86 \\
TR-MISR & 0.73 & 98.3 & 69.5 & 62.8 & 2.25 \\
PIUnet & 1.11 & 107.7 & 65.7 & 56.1 & 1.15 \\
ST-AVSR & 27.91 & 80.1 & 53.4 & 43.2 & 2.66 \\
SPS-GS (Ours) & 17.04 & \underline{75.5} & \underline{28.6} & 31.5 & 1.40 \\
\bottomrule
\end{tabular}
\end{table}

\paragraph{Efficiency.}
\Cref{tab:efficiency} reports GPU busy time measured using an NVIDIA RTX 4090. By combining single-pass encoding with Gaussian rasterization, \method achieves runtime comparable to ContinuousSR at $\times4$ (75.5 \vs 80.4\,ms), while using substantially less memory (1.40 \vs 3.44\,GiB at $\times2$). Although BurstM is slightly faster at $\times4$, it is approximately twice as slow at $\times8$ and $\times16$; BIPNet, LIIF, and GSASR also incur higher inference costs across the evaluated scales. While ContinuousSR is faster at larger scales, it uses only a single LR observation, whereas \method exploits $K{=}8$ frames. By combining input-dependent encoding with efficient Gaussian rendering, \method achieves a favorable balance between multi-frame reconstruction quality and inference efficiency.

\paragraph{Qualitative results.}
As shown in \cref{fig:qual}, single-image methods often produce blurred or distorted fine structures, such as lane markings, checkerboard patterns, and parking-lot hatching. BurstM and BIPNet recover additional details but may still exhibit structural artifacts. In contrast, \method better preserves the continuity of lane markings, the regularity of repetitive patterns, and the geometric structure of objects in these challenging examples. Additional visual comparisons, including randomly selected samples, are provided in the \appref.

\begin{table}[t]
\centering

\caption{\textbf{Results on real-world multi-frame datasets.}
cPSNR (PROBA-V and WorldStrat) and PSNR (RealBSR-RGB), in dB.
Original LR observations are used at native scales; other scales
use independently bicubic-downsampled real HR frames.
\textbf{Bold}: best; \underline{underline}: second best.}

\label{tab:realmf}
\tablefont
\setlength{\tabcolsep}{5pt}
\footnotesize
\begin{tabular}{clcccc}
\toprule
Datasets & Methods & $\times2$ & $\times4$ & $\times8$ & $\times16$ \\
\midrule
\multirow{6}{*}{\rotatebox[origin=c]{90}{PROBA-V~\cite{martens2019super}}} & Bicubic & 47.31 & 45.38 & 43.65 & 42.27 \\
 & BIPNet & 47.76 & 45.86 & 43.07 & 39.91 \\
 & HighRes-net & 48.71 & 46.57 & 43.68 & 41.28 \\
 & TR-MISR & 48.56 & 46.44 & 44.10 & 42.36 \\
 & PIUnet & \underline{49.04} & \underline{46.96} & \textbf{44.58} & \textbf{42.53} \\
 & SPS-GS (Ours) & \textbf{49.73} & \textbf{47.18} & \underline{44.53} & \underline{42.45} \\
\midrule
\multirow{6}{*}{\rotatebox[origin=c]{90}{WorldStrat~\cite{cornebise2022open}}} & Bicubic & 22.24 & 21.18 & 20.94 & 20.62 \\
 & BIPNet & \underline{24.00} & 22.46 & \underline{21.94} & \underline{21.16} \\
 & HighRes-net & 23.97 & \underline{22.47} & 21.76 & 20.64 \\
 & TR-MISR & 23.89 & 22.41 & 21.63 & 20.15 \\
 & PIUnet & 23.91 & 22.42 & 21.91 & \textbf{21.17} \\
 & SPS-GS (Ours) & \textbf{24.27} & \textbf{22.72} & \textbf{22.03} & 21.14 \\
\midrule
\multirow{6}{*}{\rotatebox[origin=c]{90}{RealBSR~\cite{wei2023towards}}} & Bicubic & 28.27 & 27.73 & 26.18 & 23.31 \\
 & BIPNet & \underline{30.02} & \underline{29.35} & 26.55 & \underline{23.47} \\
 & HighRes-net & 29.92 & 29.27 & \underline{26.64} & 23.34 \\
 & TR-MISR & 29.68 & 29.05 & 26.53 & 23.23 \\
 & PIUnet & 29.55 & 28.88 & 26.01 & 23.40 \\
 & SPS-GS (Ours) & \textbf{30.48} & \textbf{29.81} & \textbf{26.84} & \textbf{23.83} \\
\bottomrule
\end{tabular}
\end{table}

\subsection{Results on Real-World Multi-Frame Datasets}
\label{sec:realmf}
\begin{table*}[t]
\centering
\caption{Ablation of \method on AID (PSNR, dB). \cmark: sub-module used. Res./$\psi$/$\phi$/LS: residual input, phase-aware branch, shift encoding, least-squares de-aliasing; $T$/$\bmu$: \gpa iterations, center updates. Avg.: five-scale mean; $\Delta$: vs.\ the full model.}
\label{tab:ablation}
\tablefont
\setlength{\tabcolsep}{2pt}
\footnotesize
\begin{tabularx}{\textwidth}{>{\hsize=0.620\hsize\centering\arraybackslash}X>{\hsize=0.620\hsize\centering\arraybackslash}X>{\hsize=0.620\hsize\centering\arraybackslash}X>{\hsize=0.620\hsize\centering\arraybackslash}X>{\hsize=0.620\hsize\centering\arraybackslash}X>{\hsize=0.620\hsize\centering\arraybackslash}X>{\hsize=0.620\hsize\centering\arraybackslash}X>{\hsize=1.380\hsize\centering\arraybackslash}X>{\hsize=1.380\hsize\centering\arraybackslash}X>{\hsize=1.380\hsize\centering\arraybackslash}X>{\hsize=1.380\hsize\centering\arraybackslash}X>{\hsize=1.380\hsize\centering\arraybackslash}X>{\hsize=1.380\hsize\centering\arraybackslash}X>{\hsize=1.380\hsize\centering\arraybackslash}X}
\toprule
\multicolumn{4}{c}{\pcad} & \multicolumn{3}{c}{\gpa} & \multicolumn{5}{c}{PSNR$\uparrow$} & \multirow{2}{*}{Avg.} & \multirow{2}{*}{$\Delta$} \\
\cmidrule(lr){1-4}\cmidrule(lr){5-7}\cmidrule(lr){8-12}
Res. & $\psi$ & $\phi$ & LS & $T$ & Haar & $\bmu$ & $\times2$ & $\times3$ & $\times4$ & $\times6$ & $\times8$ & & \\
\midrule
 &  &  &  & 0 &  &  & 35.71 & 31.85 & 29.66 & 27.25 & 25.91 & 30.08 & $-3.09$ \\
 &  &  &  & 7 & \cmark & \cmark & 39.56 & 35.16 & 31.81 & 28.72 & 26.91 & 32.43 & $-0.74$ \\
\cmark &  &  &  & 7 & \cmark & \cmark & 39.98 & 35.31 & 32.14 & 29.00 & 27.13 & 32.71 & $-0.46$ \\
\cmark & \cmark &  &  & 7 & \cmark & \cmark & 40.02 & 35.33 & 32.18 & 29.02 & 27.17 & 32.74 & $-0.42$ \\
\cmark & \cmark & \cmark &  & 7 & \cmark & \cmark & 40.48 & 35.77 & 32.60 & 29.36 & 27.46 & 33.13 & $-0.03$ \\
\cmark & \cmark & \cmark & \cmark & 0 &  &  & 39.74 & 35.26 & 32.24 & 29.10 & 27.25 & 32.72 & $-0.45$ \\
\cmark & \cmark & \cmark & \cmark & 3 & \cmark & \cmark & 40.16 & 35.75 & 32.39 & 29.20 & 27.24 & 32.95 & $-0.22$ \\
\cmark & \cmark & \cmark & \cmark & 5 & \cmark & \cmark & 40.38 & 35.80 & 32.56 & 29.31 & 27.38 & 33.08 & $-0.08$ \\
\cmark & \cmark & \cmark & \cmark & 7 &  & \cmark & 40.44 & 35.79 & 32.60 & 29.37 & 27.42 & 33.12 & $-0.04$ \\
\cmark & \cmark & \cmark & \cmark & 7 & \cmark &  & 40.40 & 35.81 & 32.58 & 29.36 & 27.41 & 33.11 & $-0.06$ \\
\cmark & \cmark & \cmark & \cmark & 7 & \cmark & \cmark & \textbf{40.48} & \textbf{35.81} & \textbf{32.65} & \textbf{29.40} & \textbf{27.48} & \textbf{33.17} & -- \\
\bottomrule
\end{tabularx}
\end{table*}

We further evaluate on three real-world multi-frame datasets: PROBA-V~\cite{martens2019super} ($\times3$), WorldStrat~\cite{cornebise2022open} ($\times4$), and RealBSR-RGB~\cite{wei2023towards} ($\times4$). At each dataset's native scale, we use the original LR observations, while other scales are evaluated using bicubic-downsampled real HR frames, preserving inter-frame variations with synthetic degradations. Each method receives the same $K{=}8$ LR frames. All methods are trained from scratch on PROBA-V and WorldStrat and fine-tuned from their AID weights on RealBSR-RGB. For \method, inter-frame shifts are estimated via phase correlation (details in the \appref). \Cref{tab:realmf} shows that \method achieves the highest PSNR on RealBSR-RGB across all evaluated scales, outperforming the strongest baselines by up to 0.46\,dB. On PROBA-V, it performs best at $\times2$ and $\times4$ and remains competitive at higher scales. On WorldStrat, it achieves the highest cPSNR at $\times2$, $\times4$, and $\times8$, while remaining competitive at $\times16$.

\subsection{Ablation Study}
\label{sec:ablation}
\begin{figure}[t]
  \centering
  \includegraphics[width=0.9 \columnwidth]{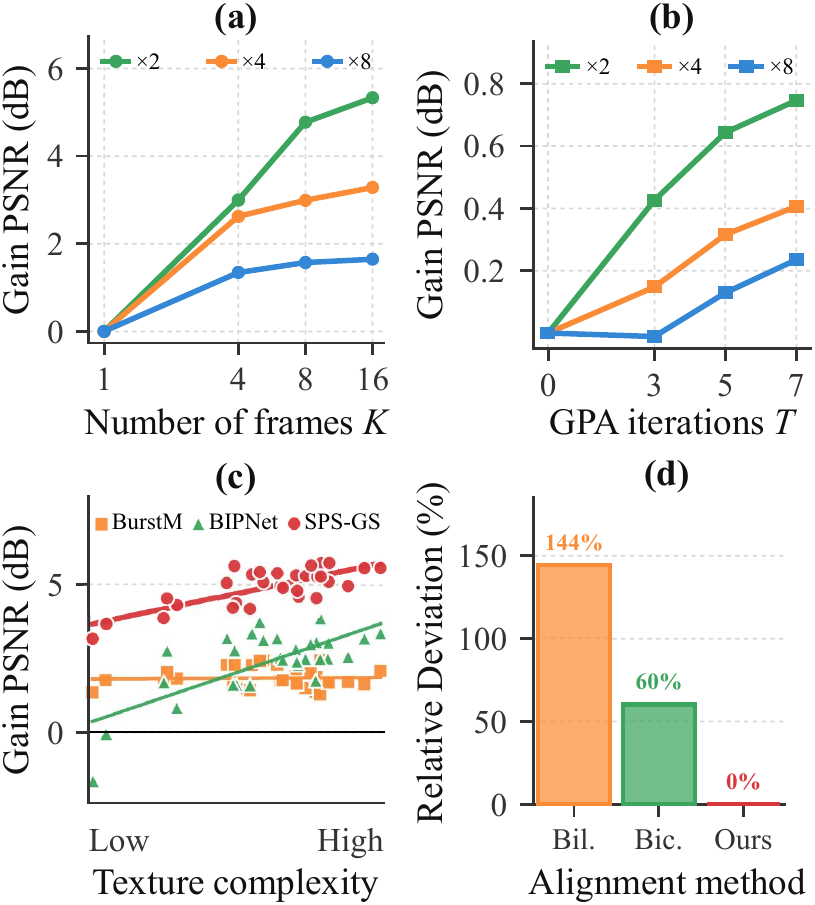}
  \caption{\textbf{Analysis of reconstruction gains.}
(a) Effect of frame count.
(b) Effect of GPA iterations.
(c) PSNR gains versus texture complexity.
(d) Relative deviations from Fourier-based alignment (reference: 0\%).}
  \label{fig:why}
  \vspace{-0.6em}
\end{figure}

\Cref{tab:ablation} removes the two main components and each of their sub-modules: the Fourier-aligned residuals, the
phase-aware branch $\psi$, the shift encoding $\phi$ and the least-squares de-aliasing of \pcad, and the iterations $T$,
the Haar sub-bands and the center updates of \gpa. \Cref{tab:repabl} replaces the 2D Gaussians by implicit decoders,
and \cref{fig:why} analyzes where the gain comes from.

\paragraph{Main-component ablation.}
As shown in \cref{tab:ablation}, removing both \pcad and \gpa reduces \method to a single-image Gaussian SR baseline, with an average PSNR of 30.08\,dB. Introducing \pcad or \gpa individually improves PSNR by up to 4.03 and 3.85\,dB, respectively. Combining both components further improves average PSNR by 0.45\,dB over \pcad alone and 0.74\,dB over \gpa alone, achieving 33.17\,dB. These results demonstrate the complementary benefits of phase-aware information aggregation and observation-guided Gaussian refinement. Similar improvements are observed in LPIPS and DISTS (\appref).

\paragraph{Sub-modules of \pcad.}
\Cref{tab:ablation} shows that Fourier-aligned residuals improve PSNR by up to 0.43\,dB with \gpa. The phase-aware branch $\psi$ provides modest gains, while shift encoding $\phi$ yields improvements of up to 0.46\,dB, highlighting the importance of shift-dependent aliasing (\cref{eq:alias}). LS estimation provides additional gains of up to 0.05\,dB, with negligible changes at $\times2$. Replacing Fourier alignment with bilinear interpolation reduces PSNR by up to 0.14\,dB, while attention pooling instead of summation decreases PSNR by up to 0.06\,dB.

\paragraph{Sub-modules of \gpa.}
\Cref{tab:ablation} examines the effect of \gpa iterations. Increasing $T$ from 0 (\pcad only) to 3, 5, and 7 progressively improves average PSNR from 32.72 to 32.95, 33.08, and 33.17\,dB, respectively, with diminishing returns. Removing the Haar sub-bands or disabling Gaussian center updates reduces average PSNR by 0.05 and 0.06\,dB, respectively, demonstrating the benefits of high-frequency error cues and spatial refinement. Additional module variants and hyperparameter analyses are provided in the \appref.

\begin{table}[t]
\centering
\caption{Representation ablation in PSNR (dB) and average inference time (ms) on AID with 2DGS and INRs.}
\label{tab:repabl}
\tablefont
\setlength{\tabcolsep}{2.6pt}
\footnotesize
\resizebox{\columnwidth}{!}{%
\begin{tabular}{lccccccc}
\toprule
Representation & $\times2$ & $\times3$ & $\times4$ & $\times6$ & $\times8$ & Avg. & Time \\
\midrule
2DGS decoder (Ours) & 40.48 & 35.81 & 32.65 & 29.40 & 27.48 & 33.17 & 69.6 \\
LIIF decoder~\cite{chen2021learning} & 40.51 & 35.82 & 32.70 & 29.42 & 27.55 & 33.20 & 488.1 \\
LTE decoder~\cite{lee2022local} & 40.50 & 35.84 & 32.73 & 29.46 & 27.52 & 33.21 & 705.9 \\
\bottomrule
\end{tabular}}
\end{table}

\paragraph{2D Gaussians \vs implicit neural representations.}
\Cref{tab:repabl} compares 2D Gaussian rasterization with the implicit decoders LIIF~\cite{chen2021learning} and LTE~\cite{lee2022local}, while keeping other components unchanged. All variants achieve comparable reconstruction accuracy, with average PSNR ranging from 33.17 to 33.21\,dB. This indicates that the reconstruction gains mainly stem from \pcad and \gpa, while 2DGS enables more efficient rendering. Specifically, 2DGS achieves approximately $7.0\times$ and $10.1\times$ faster inference than LIIF and LTE, respectively, with only a marginal PSNR difference of at most 0.04\,dB on average.

\paragraph{Analysis of reconstruction gains.}
\label{sec:analysis}
\Cref{fig:why} analyzes the factors contributing to reconstruction gains. Increasing the number of LR frames to 4, 8, and 16 yields PSNR gains of 3.0, 4.8, and 5.3\,dB at $\times2$, respectively (\cref{fig:why}a), while additional \gpa iterations provide diminishing but consistent improvements (\cref{fig:why}b). Across AID classes, the gain of \method over ContinuousSR at $\times2$ ranges from 3.2\,dB for smooth desert scenes to 5.7\,dB for texture-rich school scenes (\cref{fig:why}c), suggesting greater benefits in recovering fine structures. Moreover, \method outperforms competing methods across all 30 AID classes at $\times2$ and $\times4$. Relative to Fourier-based alignment, bilinear and bicubic interpolation exhibit normalized deviations of 144\% and 60\%, respectively (\cref{fig:why}d).

\section{Conclusion}
\label{sec:conclusion}

We presented \method, an encode-once framework for remote sensing ASSR that learns from sub-pixel-shifted LR observations synthesized from individual HR images, without requiring real multi-frame training sequences. At inference, \pcad aggregates complementary aliasing information from multiple LR observations before a single deep encoding pass, while \gpa iteratively refines Gaussian colors and centers through multi-frame back-projection to promote observation consistency. Experiments on five remote sensing benchmarks and three real-world multi-frame datasets demonstrate strong reconstruction accuracy across seen and unseen scales, with competitive inference efficiency. Future work will address radiometric variations, cloud interference, and parallax in real-world observations.

\section*{LLM Usage Statement}
We used OpenAI ChatGPT to assist with editing and polishing author-written text, refining the related-work citations, and editing author-written code. The authors take full responsibility for all submitted content.

{
    \small
    \bibliographystyle{ieeenat_fullname}
    \bibliography{main}

@inproceedings{hu2019meta,
  title={Meta-SR: A magnification-arbitrary network for super-resolution},
  author={Hu, Xuecai and Mu, Haoyuan and Zhang, Xiangyu and Wang, Zilei and Tan, Tieniu and Sun, Jian},
  booktitle={2019 IEEE/CVF Conference on Computer Vision and Pattern Recognition (CVPR)},
  pages={1575--1584},
  year={2019},
  organization={IEEE}
}

@inproceedings{chen2021learning,
  title={Learning continuous image representation with local implicit image function},
  author={Chen, Yinbo and Liu, Sifei and Wang, Xiaolong},
  booktitle={Proceedings of the IEEE/CVF conference on computer vision and pattern recognition},
  pages={8628--8638},
  year={2021}
}

@inproceedings{lee2022local,
  title={Local texture estimator for implicit representation function},
  author={Lee, Jaewon and Jin, Kyong Hwan},
  booktitle={2022 IEEE/CVF Conference on Computer Vision and Pattern Recognition (CVPR)},
  pages={1919--1928},
  year={2022},
  organization={IEEE}
}

@inproceedings{cao2023ciaosr,
  title={Ciaosr: Continuous implicit attention-in-attention network for arbitrary-scale image super-resolution},
  author={Cao, Jiezhang and Wang, Qin and Xian, Yongqin and Li, Yawei and Ni, Bingbing and Pi, Zhiming and Zhang, Kai and Zhang, Yulun and Timofte, Radu and Van Gool, Luc},
  booktitle={2023 IEEE/CVF Conference on Computer Vision and Pattern Recognition (CVPR)},
  pages={1796--1807},
  year={2023},
  organization={IEEE}
}

@inproceedings{wei2023super,
  title={Super-resolution neural operator},
  author={Wei, Min and Zhang, Xuesong},
  booktitle={2023 IEEE/CVF Conference on Computer Vision and Pattern Recognition (CVPR)},
  pages={18247--18256},
  year={2023},
  organization={IEEE}
}

@inproceedings{he2024latent,
  title={Latent modulated function for computational optimal continuous image representation},
  author={He, Zongyao and Jin, Zhi},
  booktitle={2024 IEEE/CVF Conference on Computer Vision and Pattern Recognition (CVPR)},
  pages={26026--26035},
  year={2024},
  organization={IEEE}
}

@article{becker2025thera,
  title={Thera: Aliasing-free arbitrary-scale super-resolution with neural heat fields},
  author={Becker, Alexander and Daudt, Rodrigo Caye and Narnhofer, Dominik and Peters, Torben and Metzger, Nando and Wegner, Jan Dirk and Schindler, Konrad},
  journal={arXiv preprint arXiv:2311.17643},
  year={2023}
}

@inproceedings{hu2025gaussiansr,
  title={Gaussiansr: High fidelity 2d gaussian splatting for arbitrary-scale image super-resolution},
  author={Hu, Jintong and Xia, Bin and Chen, Bin and Yang, Wenming and Zhang, Lei},
  booktitle={Proceedings of the AAAI Conference on Artificial Intelligence},
  volume={39},
  number={4},
  pages={3554--3562},
  year={2025}
}

@inproceedings{chen2025generalized,
  title={Generalized and efficient 2d gaussian splatting for arbitrary-scale super-resolution},
  author={Chen, Du and Chen, Liyi and Zhang, Zhengqiang and Zhang, Lei},
  booktitle={2025 IEEE/CVF International Conference on Computer Vision (ICCV)},
  pages={26435--26445},
  year={2025},
  organization={IEEE}
}

@inproceedings{peng2025pixel,
  title={Pixel to gaussian: Ultra-fast continuous super-resolution with 2d gaussian modeling},
  author={Peng, Long and Wu, Anran and Li, Wenbo and Zhang, Xinjie and Dai, Xueyuan and Di, Xin and Sun, Haoze and Pei, Renjing and Wang, Yang and Cao, Yang and others},
  booktitle={International Conference on Learning Representations},
  volume={2026},
  pages={92172--92199},
  year={2026}
}

@article{kerbl20233d,
  title={3d gaussian splatting for real-time radiance field rendering.},
  author={Kerbl, Bernhard and Kopanas, Georgios and Leimk{\"u}hler, Thomas and Drettakis, George and others},
  journal={ACM Trans. Graph.},
  volume={42},
  number={4},
  pages={139--1},
  year={2023}
}

@inproceedings{zhang2024gaussianimage,
  title={Gaussianimage: 1000 fps image representation and compression by 2d gaussian splatting},
  author={Zhang, Xinjie and Ge, Xingtong and Xu, Tongda and He, Dailan and Wang, Yan and Qin, Hongwei and Lu, Guo and Geng, Jing and Zhang, Jun},
  booktitle={European Conference on Computer Vision},
  pages={327--345},
  year={2024},
  organization={Springer}
}

@inproceedings{zhang2025image,
  title={Image-gs: Content-adaptive image representation via 2d gaussians},
  author={Zhang, Yunxiang and Li, Bingxuan and Kuznetsov, Alexandr and Jindal, Akshay and Diolatzis, Stavros and Chen, Kenneth and Sochenov, Anton and Kaplanyan, Anton and Sun, Qi},
  booktitle={Proceedings of the Special Interest Group on Computer Graphics and Interactive Techniques Conference Conference Papers},
  pages={1--11},
  year={2025}
}

@inproceedings{zhu2025large,
  title={Large images are Gaussians: High-quality large image representation with levels of 2D Gaussian splatting},
  author={Zhu, Lingting and Lin, Guying and Chen, Jinnan and Zhang, Xinjie and Jin, Zhenchao and Wang, Zhao and Yu, Lequan},
  booktitle={Proceedings of the AAAI Conference on Artificial Intelligence},
  volume={39},
  number={10},
  pages={10977--10985},
  year={2025}
}

@inproceedings{liang2021swinir,
  title={Swinir: Image restoration using swin transformer},
  author={Liang, Jingyun and Cao, Jiezhang and Sun, Guolei and Zhang, Kai and Van Gool, Luc and Timofte, Radu},
  booktitle={2021 IEEE/CVF international conference on computer vision workshops (ICCVW)},
  pages={1833--1844},
  year={2021},
  organization={IEEE}
}

@article{lei2021hybrid,
  title={Hybrid-scale self-similarity exploitation for remote sensing image super-resolution},
  author={Lei, Sen and Shi, Zhenwei},
  journal={IEEE Transactions on Geoscience and Remote Sensing},
  volume={60},
  pages={1--10},
  year={2021},
  publisher={IEEE}
}

@article{xiao2024ttst,
  title={TTST: A top-k token selective transformer for remote sensing image super-resolution},
  author={Xiao, Yi and Yuan, Qiangqiang and Jiang, Kui and He, Jiang and Lin, Chia-Wen and Zhang, Liangpei},
  journal={IEEE Transactions on Image Processing},
  volume={33},
  pages={738--752},
  year={2024},
  publisher={IEEE}
}

@article{chen2023continuous,
  title={Continuous remote sensing image super-resolution based on context interaction in implicit function space},
  author={Chen, Keyan and Li, Wenyuan and Lei, Sen and Chen, Jianqi and Jiang, Xiaolong and Zou, Zhengxia and Shi, Zhenwei},
  journal={IEEE Transactions on Geoscience and Remote Sensing},
  volume={61},
  pages={1--16},
  year={2023},
  publisher={IEEE}
}

@article{wu2023learning,
  author = {Wu, Hanlin and Ni, Ning and Zhang, Libao},
  title = {{Learning Dynamic Scale Awareness and Global Implicit Functions for Continuous-Scale Super-Resolution of Remote Sensing Images}},
  journal = {IEEE Transactions on Geoscience and Remote Sensing},
  volume = {61},
  pages = {1--15},
  year = {2023},
  doi = {10.1109/tgrs.2023.3240254}
}

@article{mo2026flowgs,
  title={Flow-based Gaussian Splatting for Continuous-Scale Remote Sensing Image Super-Resolution},
  author={Mo, Jiangwei and Lu, Xi and Wu, Hanlin},
  journal={arXiv preprint arXiv:2605.22147},
  year={2026}
}

@inproceedings{bhat2021deep,
  title={Deep burst super-resolution},
  author={Bhat, Goutam and Danelljan, Martin and Van Gool, Luc and Timofte, Radu},
  booktitle={2021 IEEE/CVF Conference on Computer Vision and Pattern Recognition (CVPR)},
  pages={9205--9214},
  year={2021},
  organization={IEEE}
}

@inproceedings{kang2024burstm,
  title={Burstm: Deep burst multi-scale sr using fourier space with optical flow},
  author={Kang, EungGu and Lee, Byeonghun and Im, Sunghoon and Jin, Kyong Hwan},
  booktitle={European Conference on Computer Vision},
  pages={459--477},
  year={2024},
  organization={Springer}
}

@article{martens2019super,
  title={Super-resolution of PROBA-V images using convolutional neural networks},
  author={M{\"a}rtens, Marcus and Izzo, Dario and Krzic, Andrej and Cox, Dani{\"e}l},
  journal={Astrodynamics},
  volume={3},
  number={4},
  pages={387--402},
  year={2019},
  publisher={Springer}
}

@article{deudon2020highres,
  title={Highres-net: Recursive fusion for multi-frame super-resolution of satellite imagery},
  author={Deudon, Michel and Kalaitzis, Alfredo and Goytom, Israel and Arefin, Md Rifat and Lin, Zhichao and Sankaran, Kris and Michalski, Vincent and Kahou, Samira E and Cornebise, Julien and Bengio, Yoshua},
  journal={arXiv preprint arXiv:2002.06460},
  year={2020}
}

@article{valsesia2022permutation,
  title={Permutation invariance and uncertainty in multitemporal image super-resolution},
  author={Valsesia, Diego and Magli, Enrico},
  journal={IEEE Transactions on Geoscience and Remote Sensing},
  volume={60},
  pages={1--12},
  year={2021},
  publisher={IEEE}
}

@article{an2022tr,
  title={TR-MISR: Multiimage super-resolution based on feature fusion with transformers},
  author={An, Tai and Zhang, Xin and Huo, Chunlei and Xue, Bin and Wang, Lingfeng and Pan, Chunhong},
  journal={IEEE Journal of Selected Topics in Applied Earth Observations and Remote Sensing},
  volume={15},
  pages={1373--1388},
  year={2022},
  publisher={IEEE}
}

@inproceedings{cornebise2022open,
  title={Open high-resolution satellite imagery: The worldstrat dataset--with application to super-resolution},
  author={Cornebise, Julien and Or{\v{s}}oli{\'c}, Ivan and Kalaitzis, Freddie},
  journal={Advances in Neural Information Processing Systems},
  volume={35},
  pages={25979--25991},
  year={2022}
}

@article{ni2024deformable,
  title={Deformable convolution alignment and dynamic scale-aware network for continuous-scale satellite video super-resolution},
  author={Ni, Ning and Zhang, Libao},
  journal={IEEE Transactions on Geoscience and Remote Sensing},
  volume={62},
  pages={1--17},
  year={2024},
  publisher={IEEE}
}

@article{fiete1999image,
  title={Image quality and [lambda] FN/p for remote sensing systems},
  author={Fiete, Robert D},
  journal={Optical Engineering},
  volume={38},
  number={7},
  pages={1229--1240},
  year={1999},
  publisher={SPIE}
}

@article{papoulis1977generalized,
  title={Generalized sampling expansion},
  author={Papoulis, Athanasios},
  journal={IEEE transactions on circuits and systems},
  volume={24},
  number={11},
  pages={652--654},
  year={1977},
  publisher={IEEE}
}

@inproceedings{latry2003super,
  title={Super resolution: quincunx sampling and fusion processing},
  author={Latry, Christophe and Roug{\'e}, Bernard},
  booktitle={IGARSS 2003. 2003 IEEE International Geoscience and Remote Sensing Symposium. Proceedings (IEEE Cat. No. 03CH37477)},
  volume={1},
  pages={315--317},
  year={2003},
  organization={IEEE}
}

@article{irani1991improving,
  title={Improving resolution by image registration},
  author={Irani, Michal and Peleg, Shmuel},
  journal={CVGIP: Graphical models and image processing},
  volume={53},
  number={3},
  pages={231--239},
  year={1991},
  publisher={Elsevier}
}

@inproceedings{haris2018deep,
  title={Deep back-projection networks for super-resolution},
  author={Haris, Muhammad and Shakhnarovich, Gregory and Ukita, Norimichi},
  booktitle={Proceedings of the IEEE conference on computer vision and pattern recognition},
  pages={1664--1673},
  year={2018}
}

@inproceedings{zhang2020deep,
  title={Deep unfolding network for image super-resolution},
  author={Zhang, Kai and Gool, Luc Van and Timofte, Radu},
  booktitle={Proceedings of the IEEE/CVF conference on computer vision and pattern recognition},
  pages={3217--3226},
  year={2020}
}

@article{hardie1997joint,
  title={Joint MAP registration and high-resolution image estimation using a sequence of undersampled images},
  author={Hardie, Russell C and Barnard, Kenneth J and Armstrong, Ernest E},
  journal={IEEE transactions on Image Processing},
  volume={6},
  number={12},
  pages={1621--1633},
  year={1997},
  publisher={IEEE}
}

@article{retnanto2026beyond,
  title={Beyond pretty pictures: Combined single-and multi-image super-resolution for Sentinel-2 images},
  author={Retnanto, Aditya and Le, Son and Mueller, Sebastian and Leitner, Armin and Riffler, Michael and Schindler, Konrad and Iddawela, Yohan},
  journal={arXiv preprint arXiv:2505.24799},
  year={2025}
}

@article{tarasiewicz2025graph,
  title={A graph attention network for real-world multi-image super-resolution},
  author={Tarasiewicz, Tomasz and Kawulok, Michal},
  journal={Information Fusion},
  volume={124},
  pages={103325},
  year={2025},
  publisher={Elsevier}
}

@inproceedings{lafenetre2023handheld,
  title={Handheld burst super-resolution meets multi-exposure satellite imagery},
  author={Lafenetre, Jamy and Nguyen, Ngoc Long and Facciolo, Gabriele and Eboli, Thomas},
  booktitle={2023 IEEE/CVF Conference on Computer Vision and Pattern Recognition Workshops (CVPRW)},
  pages={2056--2064},
  year={2023},
  organization={IEEE}
}

@article{ni2026satellite,
  title={Satellite Video Continuous Space-Time Super-Resolution via Mask-Based Temporal-Aware Warping and Cross-Level Frequency Integration},
  author={Ni, Ning and Zhang, Libao},
  journal={IEEE Transactions on Geoscience and Remote Sensing},
  year={2026},
  publisher={IEEE}
}

@article{xiao2024local,
  title={Local-global temporal difference learning for satellite video super-resolution},
  author={Xiao, Yi and Yuan, Qiangqiang and Jiang, Kui and Jin, Xianyu and He, Jiang and Zhang, Liangpei and Lin, Chia-wen},
  journal={IEEE Transactions on Circuits and Systems for Video Technology},
  volume={34},
  number={4},
  pages={2789--2802},
  year={2023},
  publisher={IEEE}
}

@inproceedings{di2025qmambabsr,
  title={Qmambabsr: Burst image super-resolution with query state space model},
  author={Di, Xin and Peng, Long and Xia, Peizhe and Li, Wenbo and Pei, Renjing and Cao, Yang and Wang, Yang and Zha, Zheng-Jun},
  booktitle={2025 IEEE/CVF Conference on Computer Vision and Pattern Recognition (CVPR)},
  pages={23080--23090},
  year={2025},
  organization={IEEE}
}

@article{dudhane2025burst,
  title={Burst image restoration and enhancement},
  author={Dudhane, Akshay and Zamir, Syed Waqas and Khan, Salman and Khan, Fahad Shahbaz and Yang, Ming-Husan},
  journal={IEEE transactions on pattern analysis and machine intelligence},
  volume={47},
  number={11},
  pages={9454--9467},
  year={2024},
  publisher={IEEE}
}

@inproceedings{wei2023towards,
  title={Towards real-world burst image super-resolution: Benchmark and method},
  author={Wei, Pengxu and Sun, Yujing and Guo, Xingbei and Liu, Chang and Li, Guanbin and Chen, Jie and Ji, Xiangyang and Lin, Liang},
  booktitle={2023 IEEE/CVF International Conference on Computer Vision (ICCV)},
  pages={13187--13196},
  year={2023},
  organization={IEEE}
}

@article{zhou2026tadsr,
  title={TADSR: Texture-aware dynamic Gaussian splatting for continuous-scale remote sensing image super-resolution},
  author={Zhou, Weiwei and Chang, Qianqi and Chen, Guangyong and Weng, Wuding and Gan, Min},
  journal={Optics \& Laser Technology},
  volume={203},
  pages={115812},
  year={2026},
  publisher={Elsevier}
}

@article{xiao2024ediffsr,
  title={EDiffSR: An efficient diffusion probabilistic model for remote sensing image super-resolution},
  author={Xiao, Yi and Yuan, Qiangqiang and Jiang, Kui and He, Jiang and Jin, Xianyu and Zhang, Liangpei},
  journal={IEEE Transactions on Geoscience and Remote Sensing},
  volume={62},
  pages={1--14},
  year={2023},
  publisher={IEEE}
}

@inproceedings{neuropdiff2025,
  title={Neurop-diff: Continuous remote sensing image super-resolution via neural operator diffusion},
  author={Xu, Zihao and Tang, Yuzhi and Xu, Bowen and Li, Qingquan},
  booktitle={2025 IEEE/CVF International Conference on Computer Vision (ICCV)},
  pages={12491--12501},
  year={2025},
  organization={IEEE}
}

@inproceedings{heusel2017gans,
  title={Gans trained by a two time-scale update rule converge to a local nash equilibrium},
  author={Heusel, Martin and Ramsauer, Hubert and Unterthiner, Thomas and Nessler, Bernhard and Hochreiter, Sepp},
  journal={Advances in neural information processing systems},
  volume={30},
  year={2017}
}

@inproceedings{hiif2025hierarchical,
  title={HIIF: Hierarchical encoding based implicit image function for continuous super-resolution},
  author={Jiang, Yuxuan and Kwan, Ho Man and Peng, Tianhao and Gao, Ge and Zhang, Fan and Zhu, Xiaoqing and Sole, Joel and Bull, David},
  booktitle={2025 IEEE/CVF Conference on Computer Vision and Pattern Recognition (CVPR)},
  pages={2289--2299},
  year={2025},
  organization={IEEE}
}

@article{fmsr2025frequency,
  title={Frequency-assisted mamba for remote sensing image super-resolution},
  author={Xiao, Yi and Yuan, Qiangqiang and Jiang, Kui and Chen, Yuzeng and Zhang, Qiang and Lin, Chia-Wen},
  journal={IEEE Transactions on Multimedia},
  volume={27},
  pages={1783--1796},
  year={2024},
  publisher={IEEE}
}

@inproceedings{kawar2022denoising,
  title={Denoising diffusion restoration models},
  author={Kawar, Bahjat and Elad, Michael and Ermon, Stefano and Song, Jiaming},
  journal={Advances in neural information processing systems},
  volume={35},
  pages={23593--23606},
  year={2022}
}

@inproceedings{zhang2018unreasonable,
  title={The unreasonable effectiveness of deep features as a perceptual metric},
  author={Zhang, Richard and Isola, Phillip and Efros, Alexei A and Shechtman, Eli and Wang, Oliver},
  booktitle={2018 IEEE/CVF conference on computer vision and pattern recognition},
  pages={586--595},
  year={2018},
  organization={IEEE}
}

@article{ding2022image,
  title={Image quality assessment: Unifying structure and texture similarity},
  author={Ding, Keyan and Ma, Kede and Wang, Shiqi and Simoncelli, Eero P},
  journal={IEEE transactions on pattern analysis and machine intelligence},
  volume={44},
  number={5},
  pages={2567--2581},
  year={2020},
  publisher={IEEE}
}

@article{wang2004image,
  title={Image quality assessment: from error visibility to structural similarity},
  author={Wang, Zhou and Bovik, Alan C and Sheikh, Hamid R and Simoncelli, Eero P},
  journal={IEEE transactions on image processing},
  volume={13},
  number={4},
  pages={600--612},
  year={2004},
  publisher={IEEE}
}

@article{xia2017aid,
  title={AID: A benchmark data set for performance evaluation of aerial scene classification},
  author={Xia, Gui-Song and Hu, Jingwen and Hu, Fan and Shi, Baoguang and Bai, Xiang and Zhong, Yanfei and Zhang, Liangpei and Lu, Xiaoqiang},
  journal={IEEE transactions on geoscience and remote sensing},
  volume={55},
  number={7},
  pages={3965--3981},
  year={2017},
  publisher={IEEE}
}

@inproceedings{yang2010bag,
  title={Bag-of-visual-words and spatial extensions for land-use classification},
  author={Yang, Yi and Newsam, Shawn},
  booktitle={Proceedings of the 18th SIGSPATIAL international conference on advances in geographic information systems},
  pages={270--279},
  year={2010}
}

@article{dai2011satellite,
  title={Satellite image classification via two-layer sparse coding with biased image representation},
  author={Dai, Dengxin and Yang, Wen},
  journal={IEEE Geoscience and remote sensing letters},
  volume={8},
  number={1},
  pages={173--176},
  year={2010},
  publisher={IEEE}
}

@article{zou2015deep,
  title={Deep learning based feature selection for remote sensing scene classification},
  author={Zou, Qin and Ni, Lihao and Zhang, Tong and Wang, Qian},
  journal={IEEE Geoscience and remote sensing letters},
  volume={12},
  number={11},
  pages={2321--2325},
  year={2015},
  publisher={IEEE}
}

@article{cheng2017remote,
  title={Remote sensing image scene classification: Benchmark and state of the art},
  author={Cheng, Gong and Han, Junwei and Lu, Xiaoqiang},
  journal={Proceedings of the IEEE},
  volume={105},
  number={10},
  pages={1865--1883},
  year={2017},
  publisher={IEEE}
}

@incollection{tsai1984multiframe,
  title={Multiframe image restoration and registration},
  author={Tsai, Roger Y and Huang, Thomas S},
  journal={Multiframe image restoration and registration},
  volume={1},
  pages={317--339},
  year={1984}
}

@inproceedings{shang2024arbitrary,
  title={Arbitrary-scale video super-resolution with structural and textural priors},
  author={Shang, Wei and Ren, Dongwei and Zhang, Wanying and Fang, Yuming and Zuo, Wangmeng and Ma, Kede},
  booktitle={European Conference on Computer Vision},
  pages={73--90},
  year={2024},
  organization={Springer}
}

@article{Wang2026WeakSupervision,
  title={Weak supervision makes strong details: Fine-grained object recognition in remote sensing images via regional diffusion with VLM},
  author={Wang, Liuqian and Zhang, Jing and Mi, Guangming and Zhuo, Li},
  journal={ISPRS Journal of Photogrammetry and Remote Sensing},
  volume={233},
  pages={231--246},
  year={2026},
  publisher={Elsevier}
}

@inproceedings{Ming2026HERO,
  title={Hilbert curve-encoded rotation-equivariant oriented object detector with locality-preserving spatial mapping},
  author={Ming, Qi and Wang, Liuqian and Fang, Juan and Zhao, Xudong and Xu, Yucheng and Teng, Ziyi and Zhou, Yue and Hu, Xiaoxi and Zhang, Xiaohan and Guo, Yufei},
  booktitle={Proceedings of the AAAI Conference on Artificial Intelligence},
  volume={40},
  number={10},
  pages={8071--8079},
  year={2026}
}

@article{Wang2025PGDRFNet,
  title={Position guided dynamic receptive field network: A small object detection friendly to optical and SAR images},
  author={Wang, Liuqian and Li, Jiafeng and Zhang, Jing and Zhuo, Li and Tian, Qi},
  journal={IEEE Transactions on Circuits and Systems for Video Technology},
  volume={35},
  number={8},
  pages={8265--8282},
  year={2025},
  publisher={IEEE}
}

@article{Wang2023FilterGrafting,
  author  = {Liuqian Wang and Jing Zhang and Jimiao Tian and Jiafeng Li and Li Zhuo and Qi Tian},
  title   = {{Efficient Fine-Grained Object Recognition in High-Resolution Remote Sensing Images From Knowledge Distillation to Filter Grafting}},
  journal = {{IEEE} Transactions on Geoscience and Remote Sensing},
  volume  = {61},
  pages   = {1--16},
  year    = {2023},
  doi     = {10.1109/TGRS.2023.3260883}
}

@misc{Dong2026CPGS,
  author        = {Lin Dong and Qi Ming and Liuqian Wang and Wentao Chao and Yufei Gao and Chengming Liu and Lei Shi},
  title         = {{Marrying Capacity and Precision for Efficient {2D} Gaussian Image Representation}},
  year          = {2026},
  howpublished  = {arXiv preprint arXiv:2611.13784},
  eprint        = {2611.13784},
  archivePrefix = {arXiv},
  primaryClass  = {cs.CV},
  url           = {https://arxiv.org/abs/2611.13784}
}

@misc{Gan2026GAV,
  author        = {Hao Gan and Liuqian Wang and Shuang Li and Yufei Gao and Wentao Chao and Chong Fu and Chengming Liu and Lei Shi},
  title         = {{Grounding Across Views: Cross-View Referring Reasoning for Visual Grounding}},
  year          = {2026},
  howpublished  = {arXiv preprint arXiv:2611.01074},
  eprint        = {2611.01074},
  archivePrefix = {arXiv},
  primaryClass  = {cs.CV},
  url           = {https://arxiv.org/abs/2611.01074}
}

@misc{Wang2026MOSAIC,
  author        = {Liuqian Wang and Hanyu Xie and Rongzhen Zhao and Qi Ming and Chunjie Ma and Wentao Chao and Yufei Gao and Chengming Liu and Lei Shi},
  title         = {{{MOSAIC}: Multi-Scale Object-Centric Slot Adaptation with Image-Guided Prototype Conditioning for Remote Sensing Image Super-Resolution}},
  year          = {2026},
  howpublished  = {arXiv preprint arXiv:2610.13547},
  eprint        = {2610.13547},
  archivePrefix = {arXiv},
  primaryClass  = {cs.CV},
  url           = {https://arxiv.org/abs/2610.13547}
}
}

\end{document}